\pdfoutput=1
\documentclass[11pt]{article}

\usepackage[]{acl}

\usepackage{times}
\usepackage{latexsym}

\usepackage[T1]{fontenc}
\usepackage[utf8]{inputenc}

\usepackage{microtype}

\usepackage{inconsolata}

\usepackage{graphicx}

\usepackage{newtxtext,newtxmath}
\usepackage{booktabs}
\usepackage{multirow}
\usepackage{makecell}
\usepackage{tabularx}
\usepackage{amsmath}
\usepackage{caption} 
\usepackage{siunitx}
\usepackage{graphicx}
\usepackage{adjustbox}

\usepackage{ulem}

\newcommand*\method{\texttt{InternalInspector}}

\usepackage{bm}
\usepackage{environ}

\NewEnviron{gather+}[1][1]{%
  \begin{equation}
  \scalebox{#1}{$\begin{gathered}\BODY\end{gathered}$}
  \end{equation}
}

\title{
\uline{I}nternal\uline{I}nspector I$^2$: Robust Confidence Estimation in LLMs through Internal States%}
}

\author{
 \textbf{Mohammad Beigi\textsuperscript{1}},
 \textbf{Ying Shen\textsuperscript{1}},
 \textbf{Runing Yang\textsuperscript{1}},
 \textbf{Zihao Lin\textsuperscript{1}},
\\
 \textbf{Qifan Wang\textsuperscript{2}},
 \textbf{Ankith Mohan\textsuperscript{1}},
 \textbf{Jianfeng He\textsuperscript{1}},
 \textbf{Ming Jin \textsuperscript{1}},
\\
 \textbf{Chang-Tien Lu\textsuperscript{1}},
 \textbf{Lifu Huang\textsuperscript{1}}
\\
\\
 \textsuperscript{1}Virginia Tech,
 \textsuperscript{2}Meta AI
}

\begin{document}
\maketitle

\begin{abstract}

% \Mo{I think having rl here can make the paper to be more novel for three reason in the rl version. 1) reducing the inconsistency between probing and query 2) to align the accuracy and confidence.  3) as a case study for the framework. I will remove it if you decide it is not relevant to be in the paper. This version is pure confidence estimation}

% \href{https://www.overleaf.com/1796561247tggcpydwxyjw#dca528}{link to rl version}

%\lifu{your current overleaf template is not the correct one. Double-check the website and with other students} \ying{fixed}

Despite their vast capabilities, Large Language Models (LLMs) often struggle with generating reliable outputs, frequently producing high-confidence inaccuracies known as hallucinations. 
Addressing this challenge, our research introduces \method{}, a novel framework designed to enhance confidence estimation in LLMs by leveraging contrastive learning on internal states including attention states, feed-forward states, and activation states of all layers. Unlike existing methods that primarily focus on the final activation state, \method{} conducts a comprehensive analysis across all internal states of every layer to accurately identify both correct and incorrect prediction processes. 
By benchmarking \method{} against existing confidence estimation methods across various natural language understanding and generation tasks, including factual question answering, commonsense reasoning, and reading comprehension, \method{} achieves significantly higher accuracy in aligning the estimated confidence scores with the correctness of the LLM's predictions and lower calibration error. Furthermore, \method{} excels at HaluEval, a hallucination detection benchmark, outperforming other internal-based confidence estimation methods in this task.

\end{abstract}

\section{Introduction}

Large Language Models (LLMs) have demonstrated remarkable capabilities across a wide range of tasks, from reasoning to question answering \cite{zhao2023survey, zhou2023comprehensive, Wang_2024, liang2022holistic}. 
Despite these advancements, LLMs still face significant challenges in hallucinating facts~\cite{ji2023survey, li2023halueval, 10.1145/3571730, huang2023survey} and providing robust confidence estimates for their predictions \cite{bommasani2022opportunities, kuhn2023semantic, jiang2021can}\footnote{Following \cite{kadavath2022language,jiang2021can}, we define \textit{confidence} as the probability of a model prediction being correct, differing from \textit{uncertainty} that quantifies the ambiguity in data or model lack of knowledge \cite{hu2023uncertainty}.}. 
%and providing inaccurate information \cite{ji2023survey}\lifu{add more references}. Additionally, they struggle to provide robust confidence estimates for their predictions \cite{collins2023human, }.
This results in LLMs delivering confident but incorrect information, undermining their reliability and affecting their potential applications in real-world scenarios. 
Therefore, a well-established confidence estimator is essential for users to determine when to trust the outputs of LLMs and identify hallucinations in the outputs, thereby enhancing the practicality and trustworthiness of LLMs. %Furthermore, such an estimator helps in identifying hallucinations in the outputs generated by LLMs.

Recent research suggests that LLMs exhibit a degree of self-awareness regarding the truthfulness of the generated statements \cite{kadavath2022language}. 
Studies have investigated the relationship between the accuracy of LLM outputs and the characteristics of their final activation state ~\cite{azaria-mitchell-2023-internal, burns2023discovering}. Although these findings are promising, they primarily focus on easy True/False factual question-answering tasks.
Furthermore, relying solely on the final hidden states offers a limited perspective on the intricate internal dynamics of LLMs.
The generation process within LLMs involves a sophisticated interplay of internal modules, including attention mechanisms and Feed-Forward Networks (FFNs), which are critical for shaping the generated responses. Recent studies have demonstrated that these internal modules play a crucial role in encoding and recalling the factual and linguistic knowledge essential for accurate predictions in LLMs \cite{clark2019does, kobayashi2024analyzing, modarressi-etal-2022-globenc, dar-etal-2023-analyzing, ferrando-etal-2022-measuring, modarressi-etal-2023-decompx}. It has also been shown that hallucinations in LLM outputs primarily originate from these internal modules \cite{geva2022transformer, li2023inference}. These studies further lead us to the question: \textit{can the dynamics of these internal modules potentially indicate the confidence of LLMs in their predictions?}
%This suggests that investigating the dynamics of these internal modules could potentially provide valuable guidance for estimating the confidence in generated outputs.

In response, we first provide a theoretical foundation that highlights the crucial role of internal states in robust and accurate confidence estimation. We further propose \method{} (I$^2$), a simple yet robust confidence estimation method that leverages the internal states of LLMs to assess the truthfulness of generated statements across various tasks. 
%. \method{} aims to assess the truthfulness of generated statements across various tasks, thereby enhancing the reliability and applicability of LLMs in diverse scenarios. 
% \ying{I have rewritten the following, add citation accordingly}
% \sout{Specifically, \method{} uniquely addresses the shortcomings of existing confidence estimation methods by leveraging contrastive learning to investigate the full spectrum of internal representations, particularly Multi-Head Self-Attention (MHSA) and Feed-Forward Networks (FFNs) using various architecture. }
Specifically, \method{} employs contrastive learning \cite{khosla2021supervised} upon an encoder, such as a Convolutional Neural Network \cite{o2015introduction} or a Transformer \cite{NIPS2017_7181}, to learn meaningful feature representations from the internal states of an LLM such as \texttt{LLaMA-2-7B} \cite{llama2}. A binary classifier is trained simultaneously on top of these feature representations to estimate a confidence score for each LLM prediction based on its correctness, either \textit{correct} or \textit{incorrect}.

%By utilizing this, \method{} ensures that the learned embeddings for internal states corresponding to the same types of predictions, such as all correct or all incorrect predictions, are closer, while embeddings for correct and incorrect predictions are distinctly separated. A binary classifier is then trained simultaneously on top of these embeddings to determine the confidence scores for the generated predictions.

% \sout{Additionally, METHOD distinguishes itself from other confidence estimation methods with its theoretical foundation, demonstrating why internal representations are critical for robust and accurate confidence estimation. }
% \sout{Furthermore, METHOD advances beyond previous methods that primarily rely on the final hidden state of the last layer, which fails to capture the dynamic and complex interplay of internal states that significantly influence a model's predictions. Finally, METHOD has been benchmarked against a variety of existing confidence estimation methods, including logit-based, self-evaluation, and other internal-based approaches, across a range of critical natural language generation tasks, such as factual question answering, reading comprehension, and commonsense reasoning, ensuring a comprehensive and rigorous evaluation.}

% \bigskip

We evaluate \method{} together with various existing confidence estimation methods, including logit-based, self-evaluation, and other internal-based approaches on several natural language understanding and generation tasks, including factual question answering, commonsense reasoning, and reading comprehension. Experimental results demonstrate that \method{} significantly enhances the alignment between accuracy and confidence scores, achieving up to 20.4\% improvement in accuracy and 8.9\% in Expected Calibration Error (ECE) across the various evaluation tasks. Furthermore, \method{} excels at identifying hallucinations in generated outputs, notably existing internal-based confidence estimation methods on the HaluEval benchmark~\cite{li2023halueval}. We also investigate the importance of different types of internal states in confidence estimation and showcase that attention states are particularly meaningful for tasks that require deep contextual understanding, such as open-book question answering and reading comprehension, while feed-forward states are more crucial for tasks centered on factual information, aligning with the recent research finding that the Feed-Forward Networks (FFNs) within Transformer blocks are functioning as key-value memories to encode and retrieve factual and semantic knowledge~\cite{geva2021transformer}.

In summary, our contributions are as follows:
\begin{itemize}
    \item We pioneer in establishing a theoretical foundation underscoring the importance of internal states of LLMs in confidence estimation.
    \item We propose \method{}, a simple yet effective confidence estimation method that leverages the internal states of LLMs, including the attention states, feed-forward states, and activation states.
    \item Extensive experiments demonstrate that \method{} provides robust confidence estimates and significantly outperforms existing confidence estimation methods across various natural language understanding and generation tasks.
    \item \method{} is also proven effective in recognizing hallucinations in LLM outputs, achieving significantly better performance than various baselines on HaluEval.

\end{itemize}

\section{Related Work}

% \ying{group the related work into multiple subsections, e.g., (1) understanding/interpreting the internal states/representations of LLMs. (2) Confidence Estimation for LLMs.}
% \ying{we need at least two subsections for related work}
% \ying{make it more concise (I will work on this later) }
\paragraph{Confidence Estimation for LLMs}

%In efforts to enhance the reliability of LLM outputs, various confidence estimation methods have been proposed. 

%\lifu{the following paragraphs can be merged into one or two to save space \Mo{Check at the end for checking the page limitation}}

We summarize existing confidence estimation methods for LLMs into four categories: \textbf{(1) Logit-based methods} \cite{Lin2022TeachingMT, jiang-etal-2021-know, kuhn2023semantic} utilize output probability distributions or entropy to directly measure confidence. However, they mainly reflect the probability distribution over possible tokens (vocabulary space)~\cite{Lin2022TeachingMT, si2022reexamining, tian2023just}. 
%and contrasts with human expressions of confidence, which generally reflect a belief, derived from information processing regardless of the phrasing \cite{koriat1980reasons, fischhoff1977knowing}.
\textbf{(2) Consistency-based approaches} \cite{vazhentsev-etal-2023-efficient, portillo-wightman-etal-2023-strength, wang2023selfconsistency, shi2022natural, manakul2023selfcheckgpt, agrawal2023language} evaluate confidence by measuring the agreement among different model responses, highlighting potential inconsistencies. However, these methods require effective measurement of consistency among responses which is usually challenging~\cite{xiong2024llms, jiang-etal-2021-know, Li_2022, ding2024data, kuhn2023semantic,manakul2023selfcheckgpt, zhang2023siren}. 
%face significant limitations, particularly due to the diversity of potential paraphrases and formatting variations in textual data~\cite{xiong2024llms, jiang-etal-2021-know, Li_2022, ding2024data, kuhn2023semantic} and effective measurement of consistency among responses~\cite{manakul2023selfcheckgpt, zhang2023siren}.
%. Furthermore, A non-trivial challenge in these methods is the effective measurement of consistency among responses, which remains an unresolved issue, complicating the assessment of confidence levels \cite{manakul2023selfcheckgpt, zhang2023siren}.
\textbf{(3) Self-evaluation methods} \cite{kadavath2022language, manakul2023selfcheckgpt, lin2024generating} enable models to internally assess the correctness of their answers, leveraging their introspective capability. 
%, a process compromised by the models' limited understanding and self-awareness. 
This approach often results in circular reasoning, exacerbating initial errors and leading to overconfident inaccuracies \cite{ji-etal-2023-towards, chen-etal-2023-adaptation}.
%capabilities. %Nevertheless, self-evaluation methods have inherent drawbacks. 
%They depend on the model’s ability to introspectively assess its own outputs, a process compromised by the model’s limited understanding and self-awareness. This approach often results in circular reasoning, exacerbating initial errors and leading to overconfident inaccuracies \cite{ji-etal-2023-towards, chen-etal-2023-adaptation}.
\textbf{(4) Internal-based methods} \cite{azaria2023internal, burns2022discovering} proposed training a linear classifier on the final activation state of LLMs to examine whether it can differentiate between correct and incorrect answers. 
%by mapping the activation state to probabilities. 
%Although promising, their focus on only the final activation state and reliance on linear classifiers often neglect the complex, non-linear interactions inherent in different layers of LLMs. 
\method{} falls into this category but surpasses existing methods by employing feature learning on the entire spectrum of the internal mechanism of LLMs to understand the sophisticated non-linear operational process. It generalizes effectively across various datasets and applications, offering robust confidence estimates grounded in comprehensive theoretical analysis.

% \lifu{you emphasize too much the limitations of the existing works. Just use 1-2 sentences to highlight the differences and advantages of our approach, e.g., we are employing the whole internal mechanisms instead of the final hidden state of last layer and why ours is better, and robustness etc., and use more sentences to describe the previous works. If you don't know how to write a related work paragraph, check https://arxiv.org/pdf/2402.15896v1 \Mo{Please let me know if this is ok}}

%\lifu{(1) shorten this paragraph into ~15 lines; (2) stating why this category of approach is better than previous categories and our approach falls into this category; (3) stating why our approach is better than other approaches in this category \Mo{Done}}

\paragraph{Understanding Internal States in LLMs} %\lifu{@Mo, shorten this subsection into one paragraph with around 15-20 lines} \Mo{Done}

%  \cite{clark2019does, kobayashi2024analyzing, modarressi-etal-2022-globenc, dar-etal-2023-analyzing, geva2021transformer, geva2022transformer, ferrando-etal-2022-measuring, modarressi-etal-2023-decompx} 
% Studies aimed at understanding the inner workings of transformers indicate that while attention should not be directly equated with explanation \cite{pruthi2019learning, jain2019attention, wiegreffe2019attention}, it significantly influences information processing across different layers, thereby affecting model accuracy \cite{abnar2020quantifying}. This insight into the model's operational behavior provides valuable assistance in error diagnosis and hypothesis development \cite{park2019sanvis, voita2019analyzing, vig2019visualizing, hoover-etal-2020-exbert, vashishth2019attention}.

Studies aimed at understanding the inner workings of transformers indicate that while attention should not be directly equated with explanation \cite{pruthi2019learning, jain2019attention, wiegreffe2019attention}, it provides significant insights into the model's operational behavior and helps in error diagnosis and hypothesis development \cite{park2019sanvis, voita2019analyzing, vig2019visualizing, hoover-etal-2020-exbert, vashishth2019attention}. 
Concurrently, research has shown that Feed-Forward Networks (FFNs) within Transformer blocks, functioning as key-value memories, encode and retrieve factual and semantic knowledge
%. In this system, 'keys' are human-interpretable inputs, and 'values' adjust output distributions to enhance contextual accuracy 
~\cite{geva2021transformer}. Experimental studies have established a direct correlation between modifications in FFN output distributions and subsequent token probabilities, suggesting that the model's output is crafted through cumulative updates from each layer~\cite{geva2022self}.

\section{Confidence Estimation using Internal Representations}

\subsection{Background: Transformer Architecture}

%The Transformer architecture \lifu{add reference} comprises a stack of $L$ Transformer layers, where each layer is composed of a multi-head self-attention (MHSA) sublayer followed by a feed-forward (FFN) sublayer, interconnected by residual connections that facilitate the flow of information between layers.

In this work, we primarily focused on confidence estimation for transformer-based LLMs ~\cite{vaswani2017attention}, as they have been the predominant architecture backbone of most existing frontier LLMs. Given a sequence of input tokens $x = [x_0, \cdots, x_N]$, a transformer-based language model first encodes the tokens into vectors of input representations $h^0 = [h^0_0, \cdots, h^0_N] \in \mathbb{R}^{N \times d}$ at layer $0$. The input representations are then updated through a sequence of $L$ transformer layers, where each layer is composed of a MHSA sublayer followed by a FFN sublayer, interconnected by residual connections that facilitate the flow of information between layers. 
Formally, the representation of $h^l_{i}$ of token $i$ at layer $l$ is obtained by:
\begin{align}
    h^l_{i} = h^{l-1}_{i} + a^l_{i} + m^{l}_{i},
\end{align}
where $a^l_{i}$ and $m^l_{i}$ are the outputs from the $l$-th MHSA layer and FFN sublayers, respectively.

After the transformation through $L$ layers, the representation at the final layer is projected into the vocabulary space to generate the output sequence $y$. In this work, we focus on the internal states at the final token across all layers, defined as $\theta = \{h_N^l, a_N^l, m_N^l\}_{l=1}^{L}$.
Here, $N$ represents the position of the last token in the input sequence. We select these internal states because they encapsulate the aggregation of all context information and are directly involved in producing the final predictions, making them particularly relevant for identifying the correctness of LLM's prediction.

% \ying{@Mo, need to further elaborate on how the internal states are constructed during the inference stage. Does it incorporate the internal states when generating each new token? }

% We define $\theta = \{h_N^l, a_N^l, m_N^l\}_{l=1}^{L}$ as the internal states of the last token of the LLMs, encapsulating the intricate dynamics of the model across all layers. 
% \ying{consider using symbols other than $\theta$ to represent internal states? $\theta$ is often used to represent model parameters?}

\subsection{Why Internal Representations for Confidence Estimation?}
\label{sec:theo}
% \jin{Mo and Runing, I've revised the theory section for clarity and accuracy. Pl review. \Mo{Thanks! will do}
% }

To analyze the importance of internal states $\Theta$ in assessing LLM response correctness, let $X$ and $Y$ be the input and output random variables, respectively, and $K(X)$  the oracle response (derived from expert/world knowledge) as the ground truth for a query $X$. Given an input-output pair $(X,Y)$, we define a Correctness Indicator $C(Y\mid X)$ as a binary {random} variable, taking the value 1 if $Y$ is correct given $X$, and 0 otherwise. Our confidence estimator aims to predict $C(Y\mid X)$. We assume that $C(Y\mid X)$ can be represented by a random function $S(K(X),Y)$, dependent on the oracle answer $K(X)$ and the LLM response $Y$, i.e., the Correctness Indicator is aligned with the oracle's judgment. Thus, the expected value of this function represents the Correctness Probability:
$$\mathbb{E}[S(K(X),Y)]= P(Y \text{ is correct} \mid X).$$
Let $I(\cdot ; \cdot|\cdot)$ and $H(\cdot|\cdot)$  denote the conditional mutual information/entropy, respectively. Assuming $$H(K(X)|X,Y,S(K(X),Y))\leq \epsilon,$$ where a small Residual Uncertainty $\epsilon$ indicates that $Y$, combined with the correctness indicator $S(K(X),Y)$, effectively captures most information about the oracle answer $K(X)$.\footnote{The Residual Uncertainty ($\epsilon$) Tends to be small when the Correctness Indicator is informative and the task is simple.
}

We further assume $$I(\Theta;K(X)\mid X)-I(Y;K(X)\mid X)\geq \Delta,$$ 
where the Internal Knowledge Advantage $\Delta$ quantifies the additional information about the oracle answer $K(X)$ encoded in the LLM's internal activation $\Theta$, beyond what is revealed in its output $Y$. A large $\Delta$ implies a richer internal understanding compared to the expressed output.
\footnote{Empirical evidence supports a large $\Delta$.  LLM internal states ($\Theta$) are repositories of open-world knowledge \cite{geva2021transformer, dai2022knowledge, meng2022locating}, often containing information not fully expressed in outputs. Even incorrect responses can still possess relevant knowledge internally \cite{li2023inferencetime}. Techniques like enhanced prompting \cite{wei2022chain} and self-evaluation \cite{kadavath2022language,saunders2022self, manakul2023selfcheckgpt, ren2023selfevaluation, liu2023prudent, lin2024generating} further demonstrate the ability to tap into this latent knowledge to improve accuracy, reinforcing the notion of a substantial $\Delta$.
}%, while a small or negative $\Delta$ suggests the output is a faithful representation of internal knowledge, which seems unlikely based on empirical eviden research.

Mathematically, we establish the key result:
\begin{equation}
    I(C(Y\mid X);\Theta|X,Y) \geq \Delta-\epsilon.
\end{equation}
This implies that when the internal knowledge advantage ($\Delta$) is large and the residual uncertainty ($\epsilon$) is small, the internal states ($\Theta$) provide substantial additional information about the correctness of output ($Y$) beyond what's contained in the input-output pair $(X,Y)$ alone. A detailed proof is provided in Appendix~\ref{sec:derr}.
Appendices \ref{app:quant_ass} and \ref{sec:quantitive_analysis} further explore how internal states influence performance across tasks and quantitatively analyze the impact of internal representation informativeness on confidence estimation, respectively.

\subsection{\method{}}

\paragraph{Problem Formulation} Given the dataset $\mathcal{D}=\{(x_{j}, y_{j}, \theta_{j})\}_{j=1}^M$, each instance $j$ includes an input text $x_{j}$, a generated output $y_{j}$, and the internal states $\theta_{j} = \{h^l_{N,j}, a^l_{N,j}, m^l_{N,j}\}_{l=1}^{L}$ of an LLM when generating the output $y_{j}$. Here, $N$ signifies the internal states are extracted at the last token of the input sequence.
The internal states $\theta_{j}$ include the activation states $\{h^l_{N,j}\}_{l=1}^L$, attention states $\{a^l_{N,j}\}_{l=1}^L$, and feed-forward states $\{m^l_{N,j}\}_{l=1}^L$ of the LLM across $L$ layers when processing $x_{j}$. %\lifu{make sure these terms are consistent across the whole draft \Mo{Check at the end again} \ying{pls use the present tense throughout the paper. try do it when you write new content, we may miss them when revisiting the paper}}

To effectively analyze the internal states across all layers, we stack each type of internal state along the layer dimension. 
For instance, the activation states are constructed as $h_{N,j}^{(1:L)} = [h_{N,j}^1; h_{N,j}^2; \cdots; h_{N,j}^L] \in \mathbb{R}^{L \times d}$ where $;$ denotes the concatenation along the layer dimension and $d$ is the feature dimension.
% \footnote{For clarity and readability, the subscript ${j}$ is omitted here.} 
Similarly, we form the attention states $a_{N,j}^{(1:L)}$ and the feed-forward states $m_{N,j}^{(1:L)}$ both in $\mathbb{R}^{L \times d}$.
% % \ying{TODO:better naming for these internal states?}
% We define the triplet of internal representation matrices for instance $j$ as $\theta_{j} = (h^{(1:L)}_{j}, a^{(1:L)}_{j}, m^{(1:L)}_{j})$, capturing the internal dynamics of the LLM for instance $j$.
We further construct the stacked internal states tensor, denoted as  $\theta_j = [h^{(1:L)}_{N,j}, a^{(1:L)}_{N,j}, m^{(1:L)}_{N,j}] \in \mathbb{R}^{L \times d \times 3}$ for instance $j$, capturing the entire internal dynamics of the LLM for instance $j$.

We formulate the task as learning a function $g$ that takes $\theta$ as input and outputs a confidence score $c$ indicating the correctness of $y$. Each instance $(x_j, y_j, \theta_j)$ is associated with a golden binary label $c_j$ based on whether the LLM's prediction $y_j$ is correct, where:
\[
c_j = \begin{cases} 
    1 & \text{if } y_j \text{ is correct} \\
    0 & \text{if } y_j \text{ is incorrect}.
\end{cases}
\]

%\lifu{add a paragraph to formulate the problem for InternalInspector, e.g., for each data instance, we aim to learn the InternalInspector, a function f that takes in $\theta$ as input and output a score. To train such function f, we create a label based on the correctness of the output ...} \ying{added}

\paragraph{Supervised Contrastive Learning}

\method{} employs a supervised contrastive learning framework that learns to differentiate the distinctive characteristics associated with correct and incorrect output, relying solely on the internal states. 
% Specifically, to train an effective confidence estimator, we first categorize the dataset $\mathcal{D}$ into two subsets: $\mathcal{D}_{\mathrm{correct}}$ and $\mathcal{D}_{\mathrm{wrong}}$. $\mathcal{D}_{\mathrm{correct}} = \{(x_{j}, y_{j}, \theta_{j}) \in \mathcal{D} \mid y_{j} \text{ is correct}\}$ consists of instances where the generated output is correct, while $\mathcal{D}_{\mathrm{wrong}} = \{(x_{j}, y_{j}, \theta_{j}) \in \mathcal{D} \mid y_{j} \text{ is incorrect}\}$ includes instances with incorrect predictions. 
It consists of an encoder, such as a Convolutional Neural Network (CNN)~\cite{o2015introduction} or a Transformer \cite{vaswani2017attention} (architecture detailed at Appendix \ref{App: detail}), for encoding the stacked internal states $\theta_j \in \mathbb{R}^{L \times d \times 3}$ into a compact representation $z_j = Enc(\theta_j)$. Subsequently, a multilayer perceptron (MLP) 
% \ying{@Mo, pls confirm if this is correct \Mo{Yes} } 
classifier is utilized to predict the correctness of the LLM's output $y_j$ via $\hat{c}_j = f(z_j)$. 
Aligning with the problem formulation, the overall function $g$, which maps the internal states to the confidence scores, is defined as $\hat{c}_j = g(\theta_j) = f(Enc(\theta_j))$.

\begin{figure}[t]
    \centering
        \includegraphics[width=\linewidth]{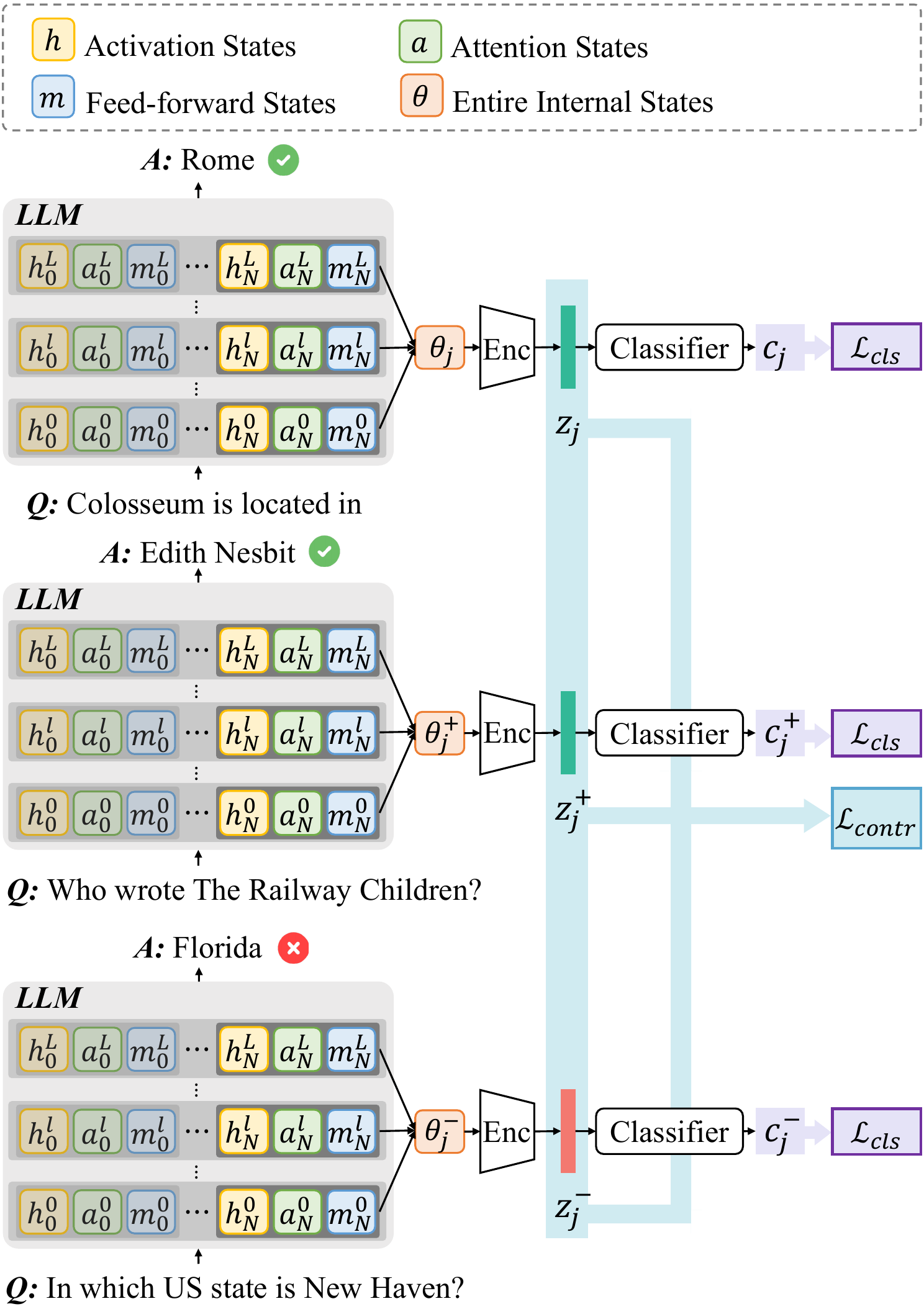}%0.95\linewidth
        \vspace{-2mm}
    \caption{\textbf{Overview of our proposed \method{}.} \method{} takes in the internal states at the final token across all layers, denoted as $\theta = \{h_N^l, a_N^l, m_N^l\}_{l=1}^{L}$, as input and outputs a confidence score $c$ indicating the correctness of the LLM's prediction.}
    \label{fig:overview}
    \vspace{-4mm}
\end{figure}

\method{} employs a combination of contrastive loss~\cite{chen2020simple} and classification loss to learn fine-grained differences in the internal states that correlate with output correctness.
For the contrastive loss, we first organize mini-batches by selecting an anchor embedding $z_j$ from the dataset. 
For each anchor $z_j$, we randomly sample one positive embedding $z_j^+$ from the set $Z^+_j = \{z_e \mid c_e = c_j \}$ ensuring that both the anchor and the positive embedding correspond to predictions with the same correctness, i.e., $c_j = c_e$. Additionally, we also sample $E$ negative embeddings $z_j^- \in Z_j^-$, where $Z_j^- = \{z_e \in Z \mid c_e \neq c_j \}$, representing the set of embeddings whose associated predictions $y_e$ differ in correctness from that of the anchor, i.e., $c_e \neq c_j$. 
Then, the contrastive loss is defined as:
\begin{align}
    \mathcal{L}_{\textrm{contr}} = - \sum_{j=1}^M \log \frac{\exp\left(z_j \cdot z_j^+ / \tau\right)}{\sum_{z_j^- \in Z_j^-} \exp\left(z_j \cdot z_j^- / \tau\right)},
\end{align}
where $\tau \in \mathbb{R}^+$ is a scalar temperature parameter.

For classification, a cross-entropy loss is used to directly optimize the model’s ability to classify the embeddings correctly:
\begin{gather+}[0.86]
\mathcal{L}_{\textrm{cls}} = - \frac{1}{M} \sum_{j=1}^M \left( c_j \log(\hat{c}_j) + (1 - c_j) \log(1 - \hat{c}_j)  \right),
\end{gather+}
where $c_j$ denotes the golden binary label of the output $y_j$.

The overall training objective is the combination of the contrastive loss and the classification loss, denoted as $\mathcal{L} = \mathcal{L}_{\textrm{contr}} + \mathcal{L}_{\textrm{cls}}$. 
The combined supervised contrastive loss empowers \method{} to effectively discern the nuances within the internal states that differentiate correct from incorrect LLM predictions, thus facilitating \method{} to accurately predict the confidence of the LLM predictions based solely on internal states.

\section{Experimental Setting}

\subsection{Tasks and Datasets}

We evaluate \method{} on one of the most popular autoregressive decoder-only open-source large language models, \texttt{LLaMA-2-7B} \cite{llama2}, on three critical tasks and datasets. For factual closed-book QA, we utilize \textbf{TriviaQA}~\cite{2017arXivtriviaqa} and \textbf{MMLU}~\cite{hendrycks2021measuring}. For commonsense reasoning, we employ \textbf{CommonsenseQA}~\cite{talmor-etal-2019-commonsenseqa} and \textbf{BoolQA}~\cite{clark2019boolq}. For reading comprehension, we utilize \textbf{SQuAD}~\cite{rajpurkar-etal-2016-squad} and \textbf{OpenBookQA}~\cite{OpenBookQA2018}.
Additionally, we also evaluate the capability of \method{} in detecting hallucinations on \textbf{HaluEval} benchmark \cite{li2023halueval}.%, designed specifically to assess hallucination phenomena in LLMs.

\subsection{Baselines}

To effectively evaluate the effectiveness of \method{}, we benchmark it against four distinct types of baseline methods:

\vspace{-2mm}
\paragraph{Logit-Based:} Following \cite{jiang-etal-2021-know}, the logit-based method utilizes the log probability derived from the output logits as a metric for confidence estimation, under the assumption that higher log probabilities suggest greater confidence. 

% \lifu{if you follow any specific implementation, you should explicitly refer it; otherwise, need to describe the specific implementation of this baseline} \Mo{They conduct the same method, so I just kept the most famous paper}

\vspace{-2mm}
\paragraph{Self-Evaluation:} Following \cite{kadavath2022language}, the Self-Evaluation method initiates a self-assessment phase after generating an answer. After the model generates an answer \(Y\), it feeds both the question \(X\) and the generated answer \(Y\) back to the model and asks whether the answer is true or false for the question. The confidence is then estimated as the probability of the generated response `True' \(P(\text{True}|X,Y)\). %\ying{what about cases when the model generates "False"? how is the confidence estimated in this case? Are we using the exact instructions in one specific paper? If so, just refer to that paper.\Mo{It is multu-choice selection, we have the logit for both true and false, we use the softmax and return the P(True). P(false) = 1-P(True)}} 

\vspace{-2mm}
\paragraph{Temperature Scaling:} %Temperature Scaling~\cite{DBLP:journals/corr/abs-2003-07892} is a widely-used post-hoc calibration method on large language models aimed to better align model confidence with actual correctness. 
Following~\cite{DBLP:journals/corr/abs-2003-07892}, Temperature Scaling adjusts the scale of logits using a scalar hyperparameter \( T \) before the softmax operation, modifying the sharpness of the probability distribution.
 
%\lifu{check the previous two baseline method descriptions and use 2-3 sentences to describe how this method works \Mo{Fixed}}

% \paragraph{Temperature Scaling:} Temperature Scaling~\cite{DBLP:journals/corr/abs-2003-07892, jiang-etal-2021-know, Lin2022TeachingMT, si2022reexamining} is one of the widely used post-hoc calibration method, aiming to better align the model’s confidence with actual answer correctness. In this method, a temperature scaling parameter, denoted as \( T \), is introduced and fine-tuned on a calibration dataset. This adjustment involves dividing the logits (pre-softmax activations) by \( T \) prior to the softmax operation. Given a logit vector \( \mathbf{z} \), the recalibrated probabilities \( \mathbf{p'} \) are expressed as:

% \begin{equation}
% \mathbf{p'} = \text{softmax}\left(\frac{\mathbf{z}}{T}\right)
% \end{equation}

% In this context, \( T > 0 \) acts as a scalar. A \( T \) of 1 retains the model's original output. When \( T \) surpasses 1, the resulting probabilities tend to be more uniform, whereas \( T \) values below 1 produce sharper, more confident predictions. 

\vspace{-2mm}
\paragraph{Last Hidden States:} We employ Contrast-Consistent Search (\textbf{CSS}) \cite{burns2023discovering}, which involves training a linear classifier on the final hidden states of statements rephrased in both positive and negative formats, and \textbf{SAPLMA}~\cite{azaria-mitchell-2023-internal} that transforms an initial statement into a true/false question and employs a classifier on the final hidden state to map it into the confidence.
%\lifu{leverage is a very vague word. Make it more specific here as people need to at least understand how these two methods work, so they can understand the difference between our approach and these baselines \Mo{Is it better?}} 
% We employ two prominent methods as baselines: \textbf{CCS}~\cite{burns2023discovering} and  \textbf{SAPLMA}~\cite{azaria2023internal}.
% The CCS approach, applied in a 0-shot configuration \ying{does CCS has the concept of using 0-shot/multi-shot? I do not seem to find CCS multi-shot in their paper?}, and SAPLMA, which concentrates on the final hidden states of models, are considered state-of-the-art in probing methodologies for confidence estimation. 
% \sout{These techniques are crucial benchmarks for evaluating the effectiveness of our framework since they represent the most advanced methods available in this field.}

\begin{table*}[h]
\begin{adjustbox}{width=0.72\width,center}
\begin{tabular}{@{}lcccccccccccc@{}}
\toprule
\textbf{Model} & \multicolumn{4}{c}{\textbf{Factual QA}} & \multicolumn{4}{c}{\textbf{Commonsense Reasoning}} & \multicolumn{4}{c}{\textbf{Reading Comprehension}} \\
\cmidrule(r){2-13}
& \multicolumn{2}{c}{TriviaQA} & \multicolumn{2}{c}{MMLU} & \multicolumn{2}{c}{CommonsenseQA} & \multicolumn{2}{c}{BoolQA} & \multicolumn{2}{c}{SQuAD} & \multicolumn{2}{c}{OpenBookQA} \\
\cmidrule(r){2-3} \cmidrule(r){4-5} \cmidrule(r){6-7} \cmidrule(r){8-9} \cmidrule(r){10-11} \cmidrule(r){12-13}
& \textbf{ACC} $\uparrow$ & \textbf{ECE} $\downarrow$ & \textbf{ACC} $\uparrow$ & \textbf{ECE} $\downarrow$ & \textbf{ACC} $\uparrow$ & \textbf{ECE} $\downarrow$ & \textbf{ACC} $\uparrow$ & \textbf{ECE} $\downarrow$ & \textbf{ACC} $\uparrow$ & \textbf{ECE} $\downarrow$ & \textbf{ACC} $\uparrow$ & \textbf{ECE} $\downarrow$ \\
\midrule
\multicolumn{13}{l}{\textit{Baseline Models}} \\
\midrule
Logit-Based & 0.453 & 0.202 & 0.416 & 0.243 & 0.576 & 0.143 & 0.532 & 0.254 & 0.522 & 0.292 & 0.464 & 0.301 \\
Temperature Scaling & 0.543 & 0.181 & 0.546 & 0.154 & 0.667 & 0.120 & 0.512 & 0.212 & 0.612 & 0.178 & 0.594 & 0.287 \\
Self-Evaluation (3-shot) & 0.307 & 0.465 & 0.374 & 0.412 & 0.312 & 0.441 & 0.282 & 0.590 & 0.309 & 0.621 & 0.368 & 0.492 \\
CSS & 0.552 &0.283&0.515&0.245&0.501&0.235&0.568&0.191&0.581&0.243&0.502&0.232 \\
SAPLMA &0.596&0.163&0.606&0.148&0.575&0.123&0.591&0.193&0.617&0.126&0.609&0.157 \\
\cmidrule(r){1-13}
\multicolumn{13}{l}{\textit{Our Models}} \\
\midrule
\method{}$_{\text{TF}}$ &\textbf{0.769}&0.081&\textbf{0.829}&\textbf{0.051}&0.742&0.102&\textbf{0.827}&0.099&0.767&0.078&0.723& 0.102\\
\method{}$_{\text{CNN}}$ &0.751&\textbf{0.073}&0.815&0.054&\textbf{0.763}&\textbf{0.097}&0.812&\textbf{0.083}&\textbf{0.807}&\textbf{0.051}&\textbf{0.791}&\textbf{0.098} \\ 
 \;  w/o Contastive loss & 0.627 &0.142&0.641&0.111&0.603&0.168&0.618&0.199&0.660&0.153&0.615& 0.171\\
% \method{}$_{\text{CNN}}$ w/o $\mathcal{L}_{\textrm{contr}}$ & 0.611 & 0.080 & 0.634 & 0.097 & 0.602 & 0.111 & 0.612 & 0.116 & 0.692 & 0.080 & 0.614 & 0.108 \\
% \method{}$_{\text{TF}}$ w/o $\mathcal{L}_{\textrm{contr}}$ & 0.643 & 0.076 & 0.676 & 0.082 & 0.636 & 0.118 & 0.659 & 0.124 & 0.637 & 0.085 & 0.657 & 0.115 \\
\bottomrule
\end{tabular}
\end{adjustbox}
\vspace{-2mm}
\caption{Comparison with baseline confidence estimation methods. Best results are highlighted in \textbf{bold}.
% \ying{the last two lines in ours in the ones without the encoder architecture? \Mo{Yes}}
}
\vspace{-4mm}
\label{tab:main_result}
\end{table*}

\subsection{Evaluation Metrics}

% \sout{We employ two primary evaluation metrics: accuracy and Expected Calibration Error (ECE). These metrics provide insights into both the performance and the reliability of the confidence score estimated by our models.}

We assess the performance of confidence estimation using two primary metrics: Accuracy and Expected Calibration Error (ECE)~\cite{Guo2017OnCO}.
\vspace{-2mm}
\paragraph{Accuracy}
This metric measures the proportion of instances where the correctness of the LLM's predictions aligns with the estimated confidence. Specifically, an output of the LLM is considered correct if its estimated confidence score exceeds a predefined threshold and incorrect if it falls below. Following~\cite{burns2023discovering, azaria-mitchell-2023-internal, li2023halueval}, we set this threshold at $0.5$ throughout our experiments, unless stated otherwise.
% \ying{what is the threshold used in our experiments? 0.5 across all tasks? following which reference? \Mo{Yes} \cite{burns2023discovering, azaria-mitchell-2023-internal, li2023halueval}}

\vspace{-2mm}
\paragraph{Expected Calibration Error (ECE)}

ECE~\cite{Guo2017OnCO} quantifies the calibration performance of the models. It is defined as:
\begin{equation}
\label{eq: ECE}
\text{ECE} = \sum_{m=1}^{M} \frac{|B_m|}{n} \left| \text{acc}({B_m}) - \text{conf}({B_m}) \right|,
\end{equation}
where $n$ is the total number of samples, $M$ is the number of bins, $B_m$ denotes the $m$-th bin containing samples with confidences falling within $\left(\frac{m-1}{M}, \frac{m}{M}\right]$, and $|B_m|$ is the number of samples in the $m$-th bin. Following \cite{desai2020calibration, kadavath2022language}, we use $M=10$ bins. $\operatorname{acc}(B_m)$ and $\operatorname{conf}(B_m)$ denote the average accuracy and confidence of the samples within $B_m$, respectively. 

\section{Results and Discussion}

\subsection{Main Results}

% \sout{Table \ref{tab:main_result} presents the performance comparison among different confidence estimation methods across multiple datasets, using accuracy (ACC) and Expected Calibration Error (ECE) as evaluation metrics.
% As illustrated, \method{} consistently outperforms all baseline methods across all evaluated tasks and datasets in both accuracy and ECE, indicating more reliable confidence in its predictions. 
% }

Table~\ref{tab:main_result} shows the performance comparison between \method{} and baseline confidence estimation methods on various tasks and datasets.  As we can see, \method{} consistently outperforms all baseline methods in terms of accuracy (ACC $\uparrow$) and Expected Calibration Error (ECE $\downarrow$), demonstrating superior performance across all evaluated tasks and datasets.
Specifically, \method{} achieves significant improvements over the highest-performing baseline, including an average 19.8\% increase in accuracy and a 8.95\% reduction in ECE for Factual QA. In Commonsense Reasoning, the improvements are 20.4\% in accuracy and 6.8\% in ECE. For Reading Comprehension, \method{} enhances accuracy by 18.6 \% and lowers ECE by 6.7\%. 

Additionally, we explore two different architectures for the encoder in \method{}: a Transformer-based encoder (\method{}$_{\text{TF}}$) and a CNN-based encoder (\method{}$_{\text{CNN}}$), as presented in Table~\ref{tab:main_result}. 
In general, \method{}$_{\text{CNN}}$ outperforms \method{}$_{\text{TF}}$ across most datasets. This superior performance is likely to be attributed to CNN's adeptness at capturing the local structure of internal states, thereby providing more effective representations for confidence estimation. The discussions of \method{} in the following sections are all based on \method{}$_{\text{CNN}}$.

Moreover, we conduct an ablation study where \method{}$_{\text{CNN}}$ is trained without the contrastive loss. This results in a notable performance decrease across all datasets, underscoring the critical role of contrastive loss in enhancing the model's effectiveness.

% To show that this component is crucial, we conducted an ablation study where \method{} was trained using the same architecture but with classification loss alone, omitting the contrastive learning module. This setup isolates the impact of contrastive learning by comparing it directly against a baseline that relies solely on classification loss. The result is detailed in the Table \ref{tab:config}. 

%\lifu{so far, there is no mention of Transformer-based encoder v.s. CNN-based encoder. You can probably add a sentence here and use one sentence to explain why CNN is better than Transformer} \ying{added}

% These statistics illustrate \method{}'s robust capability in providing reliable confidence estimations across various domains.

\begin{figure}[t]
    \centering
        \includegraphics[width=\linewidth]{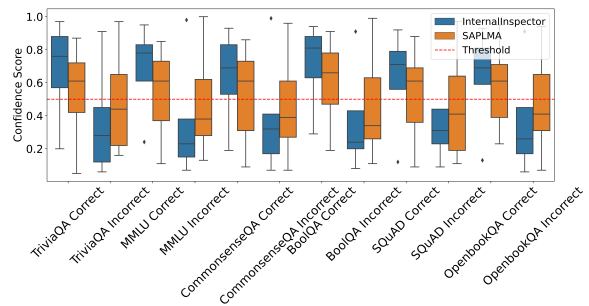}%0.95\linewidth
        \vspace{-2mm}
    \caption{Comparative Distribution of Confidence Scores. Each boxplot indicates the interquartile range of confidence scores. The dashed red line represents the decision threshold at $0.5$. 
    }
    \label{fig:dist}
    \vspace{-1em}
\end{figure}

\paragraph{Distribution of Confidence Scores}

We further examine the distribution of the estimated confidence scores from \method{} compared to those of SAPLMA, the highest-performing baseline. As depicted in Figure \ref{fig:dist}, there is a clear separation in the confidence score distributions between correct and incorrect predictions across various datasets for \method{}.
\method{} reliably maintains higher confidence scores for correct answers and lower for incorrect ones, compared to SAPLMA.
For \method{}, both the interquartile ranges and medians for correct answers consistently exceed the threshold of $0.5$, and remain below this threshold for incorrect predictions. Although SAPLMA similarly positions medians above the threshold for correct responses and below for incorrect ones, it lacks a clear separation across the interquartile ranges, indicating less reliable performance in distinguishing between correct and incorrect predictions. An analysis of high-confidence incorrect answers is in Appendix~\ref{sec:high-conf}.

\subsection{Hallucination Detection}

% \begin{table*}[tp]
% \begin{adjustbox}{width=1\width,center}
% \renewcommand{\arraystretch}{0.7}
% \begin{tabular}{@{}lcccccc@{}}
% \toprule
% \textbf{Method} & \multicolumn{2}{c}{\textbf{QA}} & \multicolumn{2}{c}{\textbf{Dialogue}} & \multicolumn{2}{c}{\textbf{Summarization}} \\
% \cmidrule(lr){2-3} \cmidrule(lr){4-5} \cmidrule(lr){6-7}
% & \textbf{ACC} & \textbf{AUC-ROC} & \textbf{ACC} & \textbf{AUC-ROC} & \textbf{ACC} & \textbf{AUC-ROC} \\
% \midrule
% Lamma 2-7B   & 48.50 & / & 44.39 & / & 47.61 & / \\
% ChatGPT & 62.59 & / & 72.40 & / & 58.53 & / \\
% w/ Knowledge & 54.39 & / & 45.12 & / & -- & / \\
% w/ CoT      & 42.73 & / & 50.51 & / & 50.91 & / \\ 
% CSS & & / & & / & & / \\
% SAPLMA & & / & & / & & / \\
% \midrule
% \method{} (Ours) & \textbf{69.15} & / & \textbf{64.83} & / & \textbf{67.13} & / \\
% \bottomrule
% \end{tabular}
% \end{adjustbox}
% \caption{Evaluation of effectiveness of the framework in hallucination detection task}
% \ying{I think we also outperform HaluEval's reported results using ChatGPT. consider including this. I have checked the leaderboard (https://huggingface.co/spaces/hallucinations-leaderboard/leaderboard). it seems like we are sota on this? double check if we are using the same experimental setup and confirm if we are really sota.
% \Mo{I added the ChatGPT result}
% }
% \label{tab:hallucination}
% \end{table*}

\begin{table}[tp]
\begin{adjustbox}{width=0.71\width,center}
\renewcommand{\arraystretch}{0.7}
\begin{tabular}{@{}lccc@{}}
\toprule
\textbf{Method} & \textbf{QA} & \textbf{Dialouge} & \textbf{Summarization} \\
\midrule
% Zero
HaluEval (Llama2-7B)   & 0.480 & 0.443 & 0.476 \\
w/ Knowledge & 0.543 & 0.451 & -- \\
w/ CoT      & 0.427 & 0.505 & 0.509 \\ \midrule
CSS &0.562&0.418&0.481 \\
SAPLMA &0.491&0.459&0.416 \\
\midrule
\method{} (Ours) & \textbf{0.691} & \textbf{0.648} & \textbf{0.671} \\
\bottomrule

\end{tabular}
\end{adjustbox}
\caption{ Accuracy (\%) of identifying whether a model's output contains hallucinated contents.
% \lifu{change the numbers to three decimal places to be consistent with other tables \Mo{Done}}
}
\vspace{-1.5em}
\label{tab:hallucination}

\end{table}

%\lifu{make sure use the tense used in the descriptions is consistent}

To assess the effectiveness of our framework in detecting hallucinations, we apply \method{} on HaluEval \cite{li2023halueval}, a hallucination evaluation benchmark for LLMs. 
The task involves taking in a question, a corresponding answer, and an optional knowledge context and identifying whether the given answer contains non-factual or hallucinated information.

In our experiments, we employ \texttt{LLaMA-2-7B} as the LLM that processes an optional knowledge content, a question, and an answer following the instruction templates in \cite{li2023halueval}. The LLM then outputs whether the provided answer is hallucinated or not.
We then employ \method{}, which utilizes the internal states of the LLM, to generate a confidence score indicating the likelihood of the answer being hallucinated.
Specifically, we train \method{} on a 30\% subset of the HaluEval and evaluate it on the remaining test split. We apply the same training setup to CSS and SAPLMA, two baseline confidence estimation methods.
We also compared with baseline methods from HaluEval, including methods with Chain of Thought (CoT) reasoning \cite{wei2022chain} and knowledge retrieval \cite{li2023web}, which are zero-shot.
As shown in Table \ref{tab:hallucination}, \method{} significantly outperforms the confidence estimation baselines in hallucination detection, suggesting \method{}'s potential in identifying hallucinations.

\subsection{Robustness on Data Distribution Shifts}

\begin{table}[tp]
\centering
\begin{adjustbox}{width=0.71\width,center}
\renewcommand{\arraystretch}{0.9} 
 
\begin{tabular}{@{}lccc@{}} 
\toprule
\textbf{Dataset} & \textbf{In-Domain} & \textbf{Intra-Domain} & \textbf{Cross-Domain}\\
\midrule
\textbf{Factual QA} & & & \\
SciQA & 0.836 & 0.737& 0.619\\
MMLU &0.815&0.707&0.598\\
\midrule
\textbf{Commonsense} & && \\
BoolQA & 0.812 & 0.694& 0.592\\
CommonsenseQA & 0.763 &   0.673& 0.585\\
\midrule
\textbf{Reading Comp.} & &  & \\
% \textbf{Comp.} & & & \\
SQuAD & 0.807 &0.683 &0.594 \\
OpenBookQA & 0.791 & 0.698 & 0.539 \\
% \midrule
% \textbf{Math} & & & \\
% Math QA & 0.352 & 0.12 & 0.14 \\
% GSMK8 & 0.235 & 0.18 & 0.16 \\
\bottomrule
\end{tabular}
\end{adjustbox}
\caption{Robustness Across Data Distribution Shifts.
% : Performance of \method{} in \texttt{In-Domain}, \texttt{Intra-Domain}, and \texttt{Cross-Domain} Settings. 
% \ying{why are we using SciQA instead of TriviQA shown in Table 1? need to explain this}
% \Mo{We discussed, the nature of SciQA and MMLU are more similar than TriviaQA. It is more reasonable for intra-domain evaluation}
}
\vspace{-1em}
\label{tab:OOD performance}
\end{table}

% \sout{In this section, we aim to investigate the robustness of \method{} under two distinct training data shift configurations for each task. In the 'in-domain' configuration, the model is trained on one dataset for a specific task and then tested on a different dataset within the same task (e.g., for factual QA, the model is trained on MMLU and tested on SciQA). 
% % \ying{do you have any reference showing this is the "in-domain" definition in confidence estimation? b/c i think "in-domain" usually means that the data is from the same data dirstribution. }
% Conversely, in the 'out-of-domain' configuration, the model is trained on datasets from various unrelated tasks and tested on a dataset from an unseen task\lifu{make it more specific}. This method allows us to assess how well \method{} generalizes across different data distributions and tasks. }
In this section, we explore \method{}'s capability to generalize across different datasets, focusing on \texttt{Intra-Domain} and \texttt{Cross-Domain} settings. 
In the \texttt{Intra-Domain} setting, \method{} is trained on one dataset of a specific task category and then tested on another dataset from the same task category. For example, within the commonsense reasoning category, the model might be trained on CommonsenseQA and tested on BoolQA. 
Conversely, in the \texttt{Cross-Domain} setting, \method{} is tested on a dataset of a specific task type while being trained on a combination of datasets from all other task types that are distinct from the test dataset's category, exemplifying its adaptability across diverse domains. 
For example, for the \texttt{Cross-Domain} scenario involving BoolQA, \method{} is evaluated on BoolQA while being trained on a combination of datasets from SciQA, MMLU, SQuAD, and OpenBook QA, none of which are within the commonsense reasoning category of BoolQA.
Additionally, we include an \texttt{In-Domain} setting, where the model is trained and tested on the same dataset to establish a baseline for comparison.
Note that, in this experiment, for tasks categorized under FactualQA, we use SciQA~\cite{SciQA2023} and MMLU~\cite{hendrycks2021measuring} due to the similarity in the subject matter they cover, containing question and answers regarding science.

Table \ref{tab:OOD performance} showcases \method{}'s robust performance in the \texttt{Intra-Domain} scenarios. Although there is a performance decrement compared to the \texttt{In-Domain} setting, \method{} consistently outperforms other baseline methods in \texttt{In-Domain} scenario (See Table \ref{tab:main_result}). 
This strong performance in \texttt{Intra-Domain} generalization indicates that the internal states from the same task category exhibit similar patterns.
% This robustness can be attributed to the operational behavior of large language models, which exhibit similar patterns and activation flows across datasets within the same task. 
% \ying{what do you mean by this?}
%, thereby facilitating \method{} in effectively learning patterns and activation flows in the internal state of LLMs. 
% \yinge{
In \texttt{Cross-Domain} setting, we observe a larger performance drop, suggesting distinct internal states and patterns across different task categories. This observation aligns with the findings that linguistic and factual knowledge located in different layers of LLMs \cite{dai2022knowledge, tenney-etal-2019-bert, meng2022locating}, resulting in task-specific variations in internal states.
Despite these variations, \method{} still performs comparably to baseline methods that are trained and evaluated on the same dataset, indicating the efficacy of \method{}.

\section{Ablation Study}

% \subsection{Effect of Contrastive Learning}

% We argue that the superiority of \method{} in comparison to CSS and SAPLMA originates from its effective implementation of contrastive learning, a feature that both CSS and SAPLMA lack. To show that this component is crucial, we conducted an ablation study where \method{} was trained using the same architecture but with classification loss alone, omitting the contrastive learning module. This setup isolates the impact of contrastive learning by comparing it directly against a baseline that relies solely on classification loss. The result is detailed in the Table \ref{tab:config}. 

% \begin{table}[tp]
% \centering
% \setlength{\tabcolsep}{3pt} 
% \renewcommand{\arraystretch}{0.9} 
% \scalebox{0.75}{ 
% \begin{tabular}{@{}lcc@{}} 
% \toprule
% \textbf{Dataset} & \textbf{\method{}} & \textbf{w/o Contrastive Loss} \\
% \midrule
% \textbf{Factual QA} & & \\
% TriviQA &  0.751 & 0.627 \\
% MMLU & 0.815 & 0.641\\
% \midrule
% \textbf{Commonsense} & & \\
% BoolQA & 0.812 & 0.603\\
% Commonsense QA & 0.763 & 0.618 \\
% \midrule
% \textbf{Reading Comp.} & & \\
% Squad & 0.807 & 0.660 \\
% Openbook QA & 0.791 & 0.615\\
% \bottomrule

% \label{tab:config}
% \end{tabular}
% }
% \caption{Performance metrics (Accuracy) across different dataset categories and model \method{} configuration.}

% \end{table}

% demonstrate ... \Mo{Waining for the experiment on it }

\subsection{Effect of Different Types of Internal States}

\begin{table}[tp]
\begin{adjustbox}{width=0.55\width,center}
\renewcommand{\arraystretch}{0.7}
\begin{tabular}{@{}lccccccc@{}}
\toprule
\textbf{Dataset} & \textbf{Full} & \textbf{FF + Attn} & \textbf{FF + Act} & \textbf{Attn + Act} & \textbf{Attn} & \textbf{FF} & \textbf{Act} \\
\midrule
\textbf{Factual QA} & & & & & & & \\

TriviaQA & 0.751 & 0.711 & 0.724& 0.613 & 0.504 & 0.703 & 0.627 \\
MMLU & 0.815 & 0.775 & 0.767 & 0.673 & 0.535 & 0.717 & 0.641 \\ % Placeholder for HS data
\midrule
\textbf{Commonsense} & & & & & & & \\
CommonsenseQA & 0.763 & 0.692 & 0.708 & 0.619 & 0.602 & 0.634 & 0.617\\ % Placeholder for HS data
BoolQA & 0.812 & 0.684 & 0.739 & 0.638 & 0.592 & 0.693 & 0.650 \\ % Placeholder for HS data
\midrule
\textbf{Reading Comp.} & & & & & & & \\
SQuAD & 0.807 & 0.771 & 0.628 & 0.759 & 0.728 & 0.615 & 0.624\\ % Placeholder for HS data
OpenBookQA & 0.791 & 0.719 & 0.615 & 0.727 & 0.707 & 0.577 & 0.635\\ % Placeholder for HS data
\bottomrule
\end{tabular}
\end{adjustbox}

\caption{Effects of utilizing different combinations of internal states, including attention states (Attn), feed-forward states (FF), and activation states (Act). \textbf{Full} represents the use of all types of internal states.}
\label{tab:embeddings}
\vspace{-2mm}
\end{table}

In this section, we explore the impact of various internal states on \method{}'s performance, focusing on the role of attention (Attn), feed-forward states (FF), activation state (Act), and their combinations across different tasks. 
Table~\ref{tab:embeddings} demonstrates that when using only one type of internal states, feed-forward states generally prove to be the most influential for confidence estimation, except for the reading comprehension task, where the model using attention states achieves the best performance. 
This highlights the effectiveness of feed-forward states in tasks that require robust factual recall, while attention states play a crucial role in tasks that necessitate processing and prioritizing extensive text segments for comprehension.

We also explore the efficacy of various combinations of internal states. The results indicate that integrating multiple types of internal states often yields improvements over using either type of internal states alone.
Moreover, models incorporating all types of internal states consistently deliver optimal performance. This suggests that the integration of different types of internal states is necessary to effectively capture the complexities inherent in various tasks, leading to robust confidence estimation methods.
These analyses offer insights into how different types or combinations of internal states might influence model performance in various task categories, potentially informing future strategies for the optimal utilization of internal states.

\subsection{Impact of Different Layer Depths}

\begin{figure}[!t]
    \centering
        \includegraphics[width=0.9\linewidth]{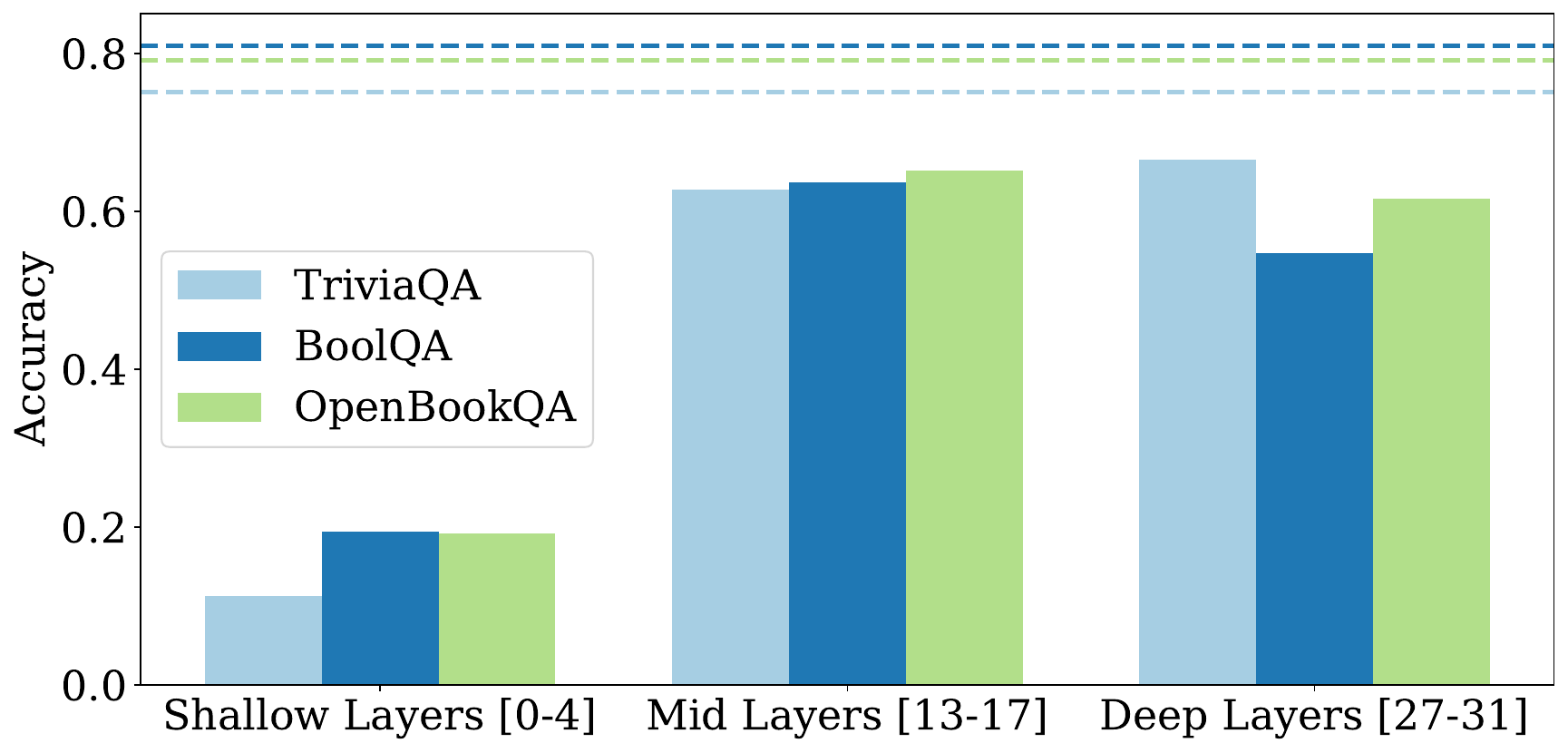}
       \vspace{-2mm}
    \caption{Impact of Internal States from Different Layer Depths.
    % : This figure illustrates the accuracy of \method{} across various tasks, showing the influence of utilizing internal states from different depths of layers.
    }
    \vspace{-4mm}
\label{fig:layer_perf}
\end{figure}

In this section, we explore the efficacy of leveraging internal states from different depths of layers within LLMs. Specifically, we analyze the performance of \method{} across diverse datasets such as TriviaQA, BoolQA, and OpenBookQA, examining how internal states from different layer depths contribute to accurate confidence estimation.

Figure~\ref{fig:layer_perf} presents the performance of \method{} when leveraging internal states from shallow (layers 0-4), middle (layers 13-17), and deep (layers 27-31) layers. The dashed horizontal line in the figure represents the baseline performance achieved when internal states from all layers are utilized.
In general, we observe that the middle layers (13-17) yield the highest performance across different tasks, suggesting that the internal states from the middle layers effectively encode features critical for assessing the correctness of model outputs. 
% This aligns with previous work that shows the middle layers of LLMs\ying{reference}
Moreover, \method{} exhibits optimal performance when internal states from all layers are utilized, underscoring the effectiveness of our current model design in leveraging internal states from all layers for confidence estimation.

\section{Conclusion}
In this work, we propose \method{}, a simple yet robust confidence estimation method utilizing the internal states of LLMs, including attention, feed-forward, and activation states across all layers. Experimental results underscore \method{}'s superior performance, which consistently outperforms baseline methods in a variety of natural language processing tasks, including factual question answering, commonsense reasoning, and reading comprehension.
% Notably, \method{} exhibits a remarkably low rate of high-confidence errors.
Further analysis shows that \method{}demonstrates strong generalization capabilities within Intra-Domain scenarios. Additionally, \method{} outperforms other internal-state-based confidence estimation methods in HaluEval, suggesting its potential in hallucination detection.

\section{Limitation}

\method{} is specifically designed to leverage the internal states of large language models (LLMs) to estimate the confidence scores of generated responses. Consequently, our proposed model cannot be applied to proprietary LLMs where these internal states are not accessible.

Moreover, in this work, we propose a simple yet effective approach for confidence estimation. While \method{} demonstrates robust performance across various tasks, we did not extensively explore complex model architectures of \method{}. Future work could delve into more advanced and complex architectures that might offer improved performance in confidence estimation.

\bibliography{main}

\appendix

%  \section{ Comparison between \dataset{} and other related datasets}\label{sec:compare}

% Table~\ref{tab:compare} compares \dataset{} with other existing datasets for multimodal entity linking.
%  \input{table/1compare_previous_dataset}

% \section{Internal Representation Figure} \label{sec:internal_representation}
% \begin{figure*}[htbp]
%     \centering
%     \includegraphics[width=1\textwidth]{emnlp2023-latex/Figures/transformer.pdf}
%   \caption{Illustration of how self-attention and FFN work in Transformer's blocks \ying{the figure is not clear enough. it seems like the three representations are color-coded, but they seem to be color-coded in different ways. try to make it clear, and highlight them} \ying{also, why specifically show the design for multi-head attention? Is this related to our method?} } 
%   \label{fig:transformers}
% \vspace{-4mm}
% \end{figure*}
\newpage
\section{Detailed Derivation in Section~\ref{sec:theo}}
\label{sec:derr}

First, we have:
\begin{align*}
    &I(\Theta;K(X)|X,Y)\\
    &=I(\Theta,X,Y;K(X))-I(X,Y;K(X))\\
    &\geq I(\Theta,X;K(X))-I(X,Y;K(X))\\
    &=\Delta.
\end{align*}
Next, observe that:
\begin{align*}
    &I(S(K(X),Y);\Theta|X,Y)\\
    &=I(S(K(X),Y),K(X);\Theta|X,Y)\\
    &\qquad\qquad-I(K(X);\Theta|X,Y,S(K(X),Y))\\
    &=I(K(X);\Theta|X,Y)\\
    &\qquad\qquad+I(S(K(X),Y);\Theta|X,Y,K(X))\\
    &\qquad\qquad-I(K(X);\Theta|X,Y,S(K(X),Y))\\
    &\geq I(K(X);\Theta|X,Y)\\
    &\qquad\qquad-I(K(X);\Theta|X,Y,S(K(X),Y))\\
    &= I(K(X);\Theta|X,Y)\\
    &\qquad\qquad-H(K(X)|X,Y,S(K(X),Y))\\
    &\qquad\qquad+H(K(X)|X,Y,S(K(X),Y),\Theta)\\
    &\geq I(K(X);\Theta|X,Y)\\
    &\qquad\qquad-H(K(X)|X,Y,S(K(X),Y))\\
    &\geq I(K(X);\Theta|X,Y)-\epsilon,
\end{align*}
where we used the fact that mutual information is nonnegative. Combining the above inequalities, we obtain the desired result.

\subsection{Quantitive assumption}
\label{app:quant_ass}
In Sec.~\ref{sec:theo}, we demonstrate the correctness $C$ and internal representations $\Theta$ are not conditionally independent on $(X,Y)$, $I(C;\Theta|X,Y) > 0$. In this section, we propose a more fine-grained theoretical model for utilizing internal representations, which use a quantified conditional mutual information $I(C;\Theta|X,Y)$ to represent the captured information between $\Theta$ and $K(\cdot)$ and introduce the reasoning confidence and knowledging confidence to explain the performance discrepancies among different datasets.  

The tested capabilities of LLMs using the QA dataset could be generally divided into two categories: \textit{knowledging and reasoning}. This decomposition is intuitively based on human cognition that knowledging determines whether the LLM is incorporated with the required knowledge to understand the question and answer the question. The reasoning capability determines whether LLM could generate the correct conclusion. Different QA datasets emphasize different aspects of these capabilities. For instance, mathematical problems focus on the LLM reasoning capability, as they require the LLM to apply logical computation to arrive at the correct answer. Closed-book QA tasks primarily assess LLM knowledging capability. Therefore we assume the confidence signal $C$ generated from the LLM could be decomposed into knowledgeing confidence $S_k$ and reasoning confidence $S_r$ and each component influences the confidence independently. For simplicity, we use a binary version of the confidence score, and the contributions of each component are described in this way:
\begin{align*}
    &P(C=1|S_r = 1) = P(C=0|S_r = 0) = \alpha\\
    &P(C=0|S_r = 1) = P(C=1|S_r = 0) = 1 - \alpha\\
    &P(C=1|S_k = 1) = P(C=0|S_k = 0) = \beta\\
    &P(C=0|S_k = 1) = P(C=1|S_k = 0) = 1- \beta\\
    &0.5 \leq \alpha, \beta \leq 1
\end{align*}
$\alpha$ represents the contribution of reasoning confidence and $\beta$ represents the contribution of knowledging confidence $\alpha$ and $\beta$ are greater than 0.5 because two confidence scores contribute positively to the correctness. Then we can derive the probability distribution of $C$ on the joint distribution of $S_r$ and $S_k$:
\begin{align*}
    &P(C = 1 | S_r = 1, S_k = 1) = 0.5(\alpha + \beta)\\
    &P(C = 1 | S_r = 1, S_k = 0) = 0.5(\alpha + 1 - \beta)\\
    &P(C = 1 | S_r = 0, S_k = 1) = 0.5(1 - \alpha + \beta)\\
    &P(C = 1 | S_r = 0, S_k = 0) = 1 - 0.5(\alpha + \beta)\\
    &P(C = 0 | S_r = 1, S_k = 1) = 1-0.5(\alpha + \beta)\\
    &P(C = 0 | S_r = 1, S_k = 0) = 0.5(1-\alpha  + \beta)\\
    &P(C = 0 | S_r = 0, S_k = 1) = 0.5(\alpha+1 - \beta)\\
    &P(C = 0 | S_r = 0, S_k = 0) = 0.5(\alpha + \beta)
\end{align*}

$I(C;\Theta|X,Y) = H(C|X,Y)-H(C|\Theta,X,Y)$. The first term $H(C|X,Y)$ depends on the data distribution of the selected dataset. To simplify, we assume the dataset $D$ contains the same number of correct and wrong answers. $P(C = 1|X,Y\in D) = P(C = 0|X,Y\in D) = \tfrac{1}{2}$. Then $H(C|X,Y) = -2*(\tfrac{1}{2})\log(\tfrac{1}{2}) = \log 2$. The second term $H(C|\Theta,X,Y) = -\sum P(\Theta,X,Y)\sum P(C|\Theta,X,Y)\log P(C|\Theta,X,Y)$, $P(C|\Theta,X,Y) = \sum P(C|S_r,S_k)P(S_r,S_k|\Theta,X,Y)$

Inspired by the interaction tensor \cite{jiang2023joint}, we consider a binarized formulation of the latent features of the internal representations and input-output pair. The three-dimension tensor $\Omega \in \{0,1\}^{M,N,T}$, where the binary label of $\Omega_{mnt} = 1$ indicates the $n^{th}$ data point contains the $t^{th}$ feature and the $m^{th}$ model learns the $t^{th}$ feature. We extend the concept of interaction tensor into our case through two modifications. 1. We do not consider multiple models but care about internal representations from multiple layers. Therefore we use the first axis with size $M$ as the hidden states from each layer. 2. We focus on two specific features of data: reasoning and knowledging. Therefore, for the third axis, we focus on two features $t_1$ related to reasoning capability and $t_2$ related to knowledging capability. Then our interaction tensor $\Omega_{mnt} = 1$ indicates the $n^{th}$ question requires the $t^{th}$ capability and the $m^{th}$ hidden states is related to the $t^{th}$ capability. In this project, we leverage all the hidden states to train the confidence estimator, therefore as long as any $m \in M$ hidden state is related to the feature required by the data point, leveraging internal representation is helpful. We define a latent embedding $\Theta_b = \{0,1\}$ of the internal representation $\Theta$ and question-answer pair $(X,Y)$. Given the $n^{th}$ data point $(X,Y)$, any $\Omega_{mnt} = 1$ indicates some hidden states capture the required capability, we set $\Theta_b =1$. Otherwise, $\Theta_b = 0$. We assume $\Theta_b$ extracts sufficient information from the question, answer, and internal representation to infer $S_r$ and $S_k$. And $S_r$ and $S_k$ are conditionally independent on $\Theta_b$. $P(S_r,S_k|\Theta_b) = P(S_r|\Theta_b)P(S_k|\Theta_b)$ 
Consider the internal representation could provide a binary signal about the reasoning confidence and knowledging confidence and the relations are described in this way:
\begin{align*}
    &P(S_r = 1|\Theta_b = 1) = P(S_r = 0 | \Theta_b = 0) = \delta\\
    &P(S_r = 0|\Theta_b = 1) = P(S_r = 1 | \Theta_b = 0) = 1- \delta\\
    &P(S_k = 1 | \Theta_b = 1) = P(S_k = 0 | \Theta_b = 0) = \epsilon\\
    &P(S_k = 0 | \Theta_b = 1) = P(S_k = 1 | \Theta_b = 0) = 1-\epsilon
\end{align*}
$\delta$ and $\epsilon$ represent how the internal representation is informative to the model reasoning confidence and knowledging confidence respectively. Then we can enumerate the probability function of $C$ conditioned on $\Theta_b$.
\begin{align*}
    &P(C=1|\Theta_b=1) \\= &P(C = 1 | S_r = 1, S_k = 1)P(S_r = 1, S_k = 1 | \Theta_b = 1)\\  + &P(C = 1 | S_r = 1, S_k = 0)P(S_r = 1, S_k = 0 | \Theta_b = 1) \\ + &P(C = 1 | S_r = 0, S_k = 1)P(S_r = 0, S_k = 1 | \Theta_b = 1) \\ + &P(C = 1 | S_r = 0, S_k = 0)P(S_r = 0, S_k = 0 | \Theta_b = 1)\\
    &=\delta\epsilon*0.5(\alpha + \beta)\\ &+ \delta(1 - \epsilon) * 0.5(\alpha + 1 - \beta)\\ &+ (1 - \delta)\epsilon * 0.5(1 - \alpha + \beta) \\&+ (1 - \delta)(1 - \epsilon) * (1 - 0.5(\alpha + \beta))
\end{align*}
\begin{align*}
    &P(C=1|\Theta_b=0) \\&= P(C = 1 | S_r = 1, S_k = 1)P(S_r = 1, S_k = 1 | \Theta_b = 0)\\  &+ P(C = 1 | S_r = 1, S_k = 0)P(S_r = 1, S_k = 0 | \Theta_b = 0) \\ &+ P(C = 1 | S_r = 0, S_k = 1)P(S_r = 0, S_k = 1 | \Theta_b = 0) \\ &+ P(C = 1 | S_r = 0, S_k = 0)P(S_r = 0, S_k = 0 | \Theta_b = 0)\\
    &=(1-\delta)(1-\epsilon)*0.5(\alpha + \beta)\\ &+ (1-\delta)\epsilon * 0.5(\alpha + 1 - \beta)\\ &+ \delta(1-\epsilon) * 0.5(1 - \alpha + \beta) \\&+  \delta\epsilon * (1 - 0.5(\alpha + \beta))
\end{align*}
\begin{align*}
    P(C=0|\Theta_b=1) &=1- P(C=1|\Theta_b=1)\\
    P(C=0|\Theta_b=0) &=1- P(C=1|\Theta_b=0)
\end{align*}
The numerical assumption about the probability relation between total confidence, reasoning and knowledging confidence, and internal representation is used to quantitatively analyze the usefulness of leveraging internal representations to predict answer correctness.
\subsection{Anslysis in Different Regimes}
\label{sec:quantitive_analysis}
To simplify the expression, 
$p_1 = P(C = 1 | \Theta_b = 1)$, $p_0 = P(C = 1 | \Theta_b = 0)$. And $P(\Theta_b = 1) = P(\Theta_b = 0) = \tfrac{1}{2}$. 
\begin{align*}
H(C|\Theta_b,X,Y) &= - [p_1 \log p_1 + (1 - p_1) \log (1 - p_1)] \\&-  [p_0 \log p_0 + (1 - p_0) \log (1 - p_0)]
\end{align*}
For a fixed LLM, the binarized latent feature $\Theta_b$ is a deterministic function of $(\Theta,X,Y)$. Therefore $I(C;\Theta_b|X,Y) \leq I(C;X,Y,\Theta|X,Y)$.
\begin{align*}
    I(C;X,Y,\Theta|X,Y) &= H(X,Y,\Theta|X,Y)\\ 
    &- H(X,Y,\Theta|X,Y,C)\\
    &= H(\Theta|X,Y) - H(\Theta|X,Y,C)\\
    &= I(C;\Theta|X,Y)
\end{align*}
In the following sections, we quantitatively analyze the relationship between the internal representations and confidence through $I(C;\Theta_b|X,Y)$ in different regimes, positive $I(C;\Theta_b|X,Y)$ implies positive $I(C;\Theta|X,Y)$.
\subsubsection{$\Theta_b$ is highly informative about the $S_r$ and $S_k$}
When the internal representation is highly informative about the $S_r$ and $S_k$. $\delta$ and $\epsilon$ could be both close to 1 or both close to 0. The former case indicates that both events often happen together and the latter case indicates one event often happens when the other event does not. Therefore, both cases indicate the strong dependence of $C$ on $S_r$ and $S_k$. When both $\delta$ and $\epsilon$ are close to 1, $p_1\approx 0.5(\alpha+\beta)$ and $p_0\approx 1-0.5(\alpha+\beta)$. When both $\delta$ and $\epsilon$ are close to 0, $p_0\approx 0.5(\alpha+\beta)$ and $p_1\approx 1-0.5(\alpha+\beta)$. Introduce $\gamma = 0.5(\alpha+\beta)\geq0.5$ for simplification. In both cases, $H(C|\Theta_b, X,Y)\approx-P(\Theta_b=1)[\gamma \log (\gamma)+(1-\gamma)\log (1-\gamma)]-P(\Theta_b=0)[\gamma \log (\gamma)+(1-\gamma)\log (1-\gamma)] \approx -\gamma \log (\gamma)+(1-\gamma)\log (1-\gamma)$. The conditional entropy is a concave function with respect to $\gamma$, which achieves maximum value at $\eta = 0.5$. $I(C; \Theta_b | X,Y) = H(C | X, Y) - H(C | \Theta_b,X,Y) \approx \log 2 - \gamma \log \gamma + (1 - \gamma) \log (1 - \gamma)$ is non-negative when $\gamma \geq \tfrac{1}{2}$
\subsubsection{$\Theta_b$ provides little information about $S_r$ and $S_k$}
When $\Theta_b$ provides little information about $S_r$ and $S_k$, $\delta$ and $\epsilon$ are close to 0.5. Then $p_1 \approx 0.5$ and $p_0 \approx 0.5$. $H(C|\Theta_b, X,Y) \approx - P(\Theta_b = 1) [0.5 \log 0.5 + 0.5 \log 0.5] - P(\Theta_b = 0) [0.5 \log 0.5 + 0.5 \log 0.5] = \log 2$. $I(C; \Theta_b | X,Y) = H(C | X, Y) - H(C | \Theta_b,X,Y) \approx \log 2-\log 2$ is minimized near to 0.

\subsubsection{$S_r$ and $S_k$ contributes to $C$} When $S_r$ and $S_k$ contributes positively to $C$, $\alpha$ and $\beta$ are close to 1. Then we get $p_1\approx\frac{\delta+\epsilon}{2}$ and $ p_0\approx1-\frac{\delta+\epsilon}{2}$. Introduce $\eta = \frac{\delta+\epsilon}{2}$ for simplification.
$H(C|\Theta_b, X,Y) \approx - P(\Theta_b = 1) [\eta \log \eta + (1 - \eta) \log (1 - \eta)] - P(\Theta_b = 0) [(1 - \eta) \log (1 - \eta) + \eta \log \eta]
= - \eta \log \eta + (1 - \eta) \log (1 - \eta)$.
The conditional entropy is a concave function with respect to $\eta$, which achieves maximum value at $\eta = 0.5$. $I(C; \Theta_b | X,Y) = H(C | X, Y) - H(C | \Theta_b,X,Y) \approx \log 2 - \eta \log \eta + (1 - \eta) \log (1 - \eta)$ is non-negative $\eta \geq \tfrac{1}{2}$ 

\subsubsection{$S_r$ and $S_k$ are not correlated to the $C$} When $S_r$ and $S_k$ are not correlated to the $C$, $\alpha$ and $\beta$ are close to 0.5. Then we get $p_0\approx0.5$ and $p_1\approx0.5$.
$H(C|\Theta_b, X,Y) \approx - P(\Theta_b = 1) [0.5 \log 0.5 + 0.5 \log 0.5] - P(\Theta_b = 0) [0.5 \log 0.5 + 0.5 \log 0.5] = \log 2$. $I(C; \Theta_b | X,Y) = H(C | X, Y) - H(C | \Theta_b,X,Y) \approx \log 2-\log 2$ is minimized near to 0.

\label{app: implementation}

\section{High-Confidence Incorrect Answers}
\label{sec:high-conf}
\begin{figure}[htpb]
    \centering
        \includegraphics[width=0.9\linewidth]{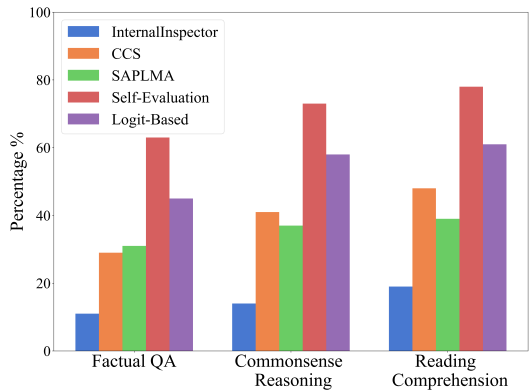}
       \vspace{1em}
    \caption{Percentage of high-confidence incorrect answers across various tasks. 
    }
    \vspace{-4mm}
\label{fig:high-confidence}
\end{figure}

We further extend our analysis to specifically focus on high-confidence incorrect answers, a critical metric for evaluating the reliability of confidence estimation methods. This analysis is crucial for identifying overconfidence in model predictions, which can have severe implications in high-stakes scenarios. 
Following the guidelines suggested in \cite{lin2022teaching, mielke-etal-2022-reducing, lin2022teaching}, we examine instances where the model, despite incorrect predictions, assigns disproportionately high confidence levels — scores above $0.8$.  Figure \ref{fig:high-confidence} compares the percentage of high-confidence incorrect predictions across various confidence estimation methods. 
The results demonstrate that \method{} maintains a significantly lower percentage of high-confidence errors across all datasets and tasks compared to other baselines. This performance underscores the enhanced calibration capability of \method{}, effectively minimizing the risk associated with overconfident misjudgments and thereby improving the model's overall reliability.

\section{Implementation Detail and Hyperparameters}
\label{App: detail}

\subsection{Encoder Architecture}
We employed the ResNet18 architecture \cite{he2016deep} to encode the internal states of the LLM. The deep residual learning framework of ResNet18 efficiently captures relationships both within and across different layers of the LLM. In designing the Transformer as an alternative encoder for our experimental setup, we implemented an 8-layer configuration without an input embedding module, utilizing a model dimensionality of \(d = 768\). Each of these layers comprises a self-attention layer and an MLP layer. Atop the feature representations obtained from either the Transformer or the CNN encoder, we train a binary classifier consisting of a multilayer perceptron (MLP) with three layers. This MLP is configured with ReLU activation to effectively process and classify the nuanced, high-dimensional encoded data. To enhance the classifier’s robustness and prevent overfitting, L2 regularization and a dropout rate of 0.1 are incorporated into the MLP. It is optimized using a learning rate of 0.001, ensuring stable and efficient learning dynamics.

\subsection{Computational Resources}

We conduct our experiments on a server equipped with four NVIDIA A100 Tensor Core GPUs. Training \method{} on various datasets is efficiently completed in under four hours using two of these GPUs.

\subsection{Data Split}

We train \method{} on the training split of the datasets and evaluate its performance on the test split.

\end{document}